\documentclass[sigconf]{acmart}
\AtBeginDocument{%
  }

\copyrightyear{2026}
\acmYear{2026}
\setcopyright{cc}
\setcctype{by-nc-nd}
\acmDOI{10.1145/3770855.3818323}
\acmConference[KDD '26]{Proceedings of the 32nd ACM SIGKDD Conference on Knowledge Discovery and Data Mining V.2}{August 09--13, 2026}{Jeju Island, Republic of Korea}
\acmBooktitle{Proceedings of the 32nd ACM SIGKDD Conference on Knowledge Discovery and Data Mining V.2 (KDD '26), August 09--13, 2026, Jeju Island, Republic of Korea}
\acmISBN{979-8-4007-2259-2/2026/08}

\usepackage{xspace}
\usepackage{multirow}
\usepackage[export]{adjustbox}
\usepackage{CJKutf8}
\usepackage{balance}
\usepackage{xcolor}
\usepackage[table]{xcolor}

\newcommand{\chinese}[1]{\begin{CJK}{UTF8}{gbsn}#1\end{CJK}}
\DeclareRobustCommand{\ourmethod}{LASEV\xspace}
\begin{document}

\title{Beyond End-to-End Video Models: An LLM-Based \\
Multi-Agent System for Educational Video Generation}


\author{Lingyong Yan}
\authornote{Both authors contributed equally.}
\affiliation{%
  \institution{Baidu Inc.}
  \city{Beijing}
  \country{China}}
\email{yanlingyong@baidu.com}

\author{Jiulong Wu}
\authornotemark[1]
\affiliation{%
  \institution{Baidu Inc.}
  \city{Beijing}
  \country{China}}
\email{wujiulong@baidu.com}

\author{Dong Xie}
\affiliation{%
  \institution{Baidu Inc.}
  \city{Beijing}
  \country{China}}
\email{xiedong04@baidu.com}

\author{Weixian Shi}
\affiliation{%
  \institution{Baidu Inc.}
  \city{Beijing}
  \country{China}}
\email{shiweixian@baidu.com}

\author{Deguo Xia}
\affiliation{%
  \institution{Baidu Inc.}
  \city{Beijing}
  \country{China}}
\email{xiadeguo@baidu.com}

\author{Jizhou Huang}
\authornote{Project lead and corresponding author.}
\affiliation{%
  \institution{Baidu Inc.}
  \city{Beijing}
  \country{China}}
\email{huangjizhou@baidu.com}

\renewcommand{\shortauthors}{Yan et al.}

\begin{abstract}
Although recent end-to-end video generation models demonstrate impressive performance in visually oriented content creation, they remain limited in scenarios that require strict logical rigor and precise knowledge representation, such as instructional and educational media.
To address this problem, we propose \ourmethod, a hierarchical LLM-based multi-agent system for generating high-quality instructional videos from educational problems.
\ourmethod formulates educational video generation as a multi-objective task that simultaneously demands correct step-by-step reasoning, pedagogically coherent narration, semantically faithful visual demonstrations, and precise audio--visual alignment.
To address the limitations of prior approaches—including low procedural fidelity, high production cost, and limited controllability—\ourmethod decomposes the generation workflow into specialized agents that collaborate through a central Orchestrating Agent, shared production state, explicit quality gates, and iterative critique mechanisms.
Specifically, the Orchestrating Agent supervises a Solution Agent for rigorous problem solving, an Illustration Agent that produces executable visualization code, and a Narration Agent for learner-oriented instructional scripts.
In addition, all outputs from the working agents are subject to semantic critique, rule-based constraints, and tool-based compilation checks.
Rather than directly synthesizing pixels, the system constructs a structured executable video script that is deterministically compiled into synchronized visuals and narration using template-driven assembly rules, enabling fully automated production without manual editing.
In large-scale deployments, \ourmethod achieves a throughput exceeding one million videos per day, delivering over a 95\% reduction in cost compared to current industry-standard approaches while maintaining a high acceptance rate.
The project is available at \url{https://robitsg.github.io/LASEV}.
\end{abstract}




\maketitle

\section{Introduction}
\label{sec:introduction}

Recent advances in large-scale generative architectures have significantly improved the capabilities of video generation models~\cite{openai2024sora,jiang2024morahighrankupdatingparameterefficient,wang2025mavis}. 
By leveraging text-conditioned learning and the reasoning abilities of large language models (LLMs), modern text-to-video models can produce visually coherent, temporally consistent, and aesthetically appealing videos~\cite{tang2025video}. 
In practice, representative models and systems, such as Sora~\cite{openai2024sora} and the Wan series~\cite{wan_etal_2025_wan}, demonstrate strong performance on open-ended and visually driven generation tasks, exhibiting impressive capabilities in motion synthesis, style consistency, and scene continuity~\cite{ghafoorian2026rehyatrecurrenthybridattention,wang2025emcontrol,wang2025lavie}. 
These advances have further enabled higher-resolution generation and coherent multi-shot narratives, supporting a wide range of visually oriented content creation scenarios~\cite{wan_etal_2025_wan,gao_etal_2025_seedance_10,seedance_etal_2025_seedance_15_pro}.

Despite their strong performance in perceptual-level synthesis, existing end-to-end models face fundamental limitations in scenarios requiring strict logical rigor and precise knowledge representation.
This limitation is particularly pronounced in educational and instructional contexts.
For example, unlike open-ended artistic generation, K--12 problem-solving videos must adhere to well-defined pedagogical workflows, necessitating accurate problem presentation, step-by-step analytical reasoning, and rigorous solution execution~\cite{anderson2001taxonomy,rosenshine2012principles,mayer2021multimedia}.
Such content demands strict global consistency across textual, symbolic, and visual modalities, as any deviation in intermediate steps or logical reasoning may severely impair learning outcomes~\cite{sweller2011cognitive}.
However, most current text-to-video models are optimized for visual realism rather than procedural fidelity. 
Since these models operate in a probabilistic pixel space, they struggle to faithfully render deterministic symbolic content—leading to common failures such as distorted equations, inconsistent diagrams, and flawed logical progressions.

Recent advances in LLMs provide a promising alternative by offering strong capabilities in knowledge reasoning, symbolic manipulation, and structured code generation~\cite{achiam2023gpt,touvron2023llama,zhao2023survey}.
These abilities provide a foundational basis for modeling complex instructional procedures and enforcing structured generation constraints. 
Furthermore, LLM-based multi-agent systems have shown immense potential in coordinating long-horizon workflows through task decomposition, role specialization, and iterative collaboration~\cite{hong2023metagpt,wu2024autogen,yan_etal_2025_beyond,tran_etal_2025_multiagent,chen2025improving,li_etal_2025_a_unified}.
By decomposing pedagogical video generation into modular and controllable stages, such systems offer a new paradigm for structured content creation.
These developments motivate our core research question: \textit{Can educational video generation workflows be systematically modeled and executed by LLM-based multi-agent systems to ensure \textbf{accurate, consistent, and procedure-aware instructional video synthesis}?}

To address this question, we formulate the instructional video generation process as a structured LLM-based multi-agent system, which combines role-specialized agent modules with a collaborative multi-agent harness for orchestration, validation, and template-based assembly.
In this system, specialized LLM-based agents are tasked with distinct roles: rigorous reasoning, executable illustration, and pedagogical narration.
We model video synthesis as a progressive script construction process, in which intermediate reasoning steps, visual specifications, and pedagogical explanations are incrementally organized into unified, machine-executable scripts with explicit structural constraints.
This modular design not only alleviates the cognitive burden on individual models but also facilitates fine-grained supervision at each stage of content generation.
Crucially, by transforming video synthesis into a controllable script generation problem, our approach enables LLMs to precisely regulate semantic structure, temporal coherence, and pedagogical consistency, ensuring rigor throughout the generated content.

Based on this formulation, we propose \textbf{\ourmethod} (\textbf{L}LM-based Multi-\textbf{A}gent \textbf{S}ystem for \textbf{E}ducational \textbf{V}ideo), an end-to-end system for high-fidelity instructional video generation.
First, \ourmethod adopts a collaborative multi-agent harness: the central Orchestrating Agent decomposes high-level instructional objectives into executable subtasks, dispatches them to role-specialized working agents, and aggregates validated outputs through a shared production state.
Under this harness, three specialized working agents operate in concert: 
(1) a \textit{Solution Agent} for rigorous problem solving, generating step-by-step analytical reasoning; 
(2) an \textit{Illustration Agent} for visual content generation, which produces executable visualization code (e.g., Python/Manim) rather than direct pixel synthesis; and 
(3) a \textit{Narration Agent} for pedagogical script generation, producing learner-friendly explanatory texts aligned with the solution.
The Orchestrating Agent systematically aggregates these validated outputs into a unified video specification, combining structured scripts, rendering code, and synchronized narration for final compilation.

Second, to ensure the output quality and global consistency, \ourmethod incorporates a \textit{heterogeneous critique mechanism} that provides multi-level verification.
Each agent's output undergoes rigorous inspection across three dimensions: \textbf{semantic rubrics} for pedagogical correctness, \textbf{tool-based execution} for functional feasibility (e.g., code compilation), and \textbf{rule-based constraints} for structural compliance (e.g., function usage and keyword matching).
When any validation criterion is violated, detailed critique feedback triggers an iterative refinement loop.
This critique--revision cycle continues until all constraints are satisfied, enabling systematic error correction and stable convergence toward high-quality instructional content.

In summary, the main contributions of this paper are as follows:
\begin{enumerate}
    \item We propose a structured LLM-based multi-agent system for instructional video generation, which decomposes complex pedagogical workflows into coordinated reasoning, visualization, and narration stages under centralized orchestration.
    
    \item We introduce a heterogeneous critique and verification mechanism that integrates semantic rubrics, tool-based execution, and rule-based constraints, ensuring both logical consistency and functional correctness.
    
    \item We establish an end-to-end automated pipeline that unifies multi-agent collaboration, structured scripting, and executable visualization code, enabling controllable, scalable, and high-fidelity instructional video synthesis.
\end{enumerate}
\section{Related Work}

\paragraph{\textbf{Multi-Agent Systems}}
The rapid development of Large Language Models (LLMs) has transformed multi-agent systems (MAS) from predefined rule-based entities into autonomous collaborators capable of solving complex problems. 
Early systems primarily utilize role-playing strategies, where agents are assigned specific personas (e.g., coder, reviewer) to decompose tasks into manageable sub-steps~\cite{hong2023metagpt,li2023camel,qian2023communicative}. 
To support more flexible interactions, subsequent research introduced conversation-based architectures. 
Platforms such as AutoGen~\cite{wu2024autogen} and AgentVerse~\cite{chen2023agentverse} enable developers to customize communication patterns, allowing agents to negotiate and refine their outputs through multi-turn dialogue~\cite{park2023generative}.
More recently, the focus has shifted towards enhancing the reasoning and planning capabilities of these agents. 
Advanced systems now incorporate ``reflective'' mechanisms, where agents self-correct their errors by analyzing past actions~\cite{shinn2023reflexion,bo2024reflective}. 
By leveraging models trained with reinforcement learning for extensive chain-of-thought reasoning, such as DeepSeek-R1 ~\cite{guo2025deepseek} and OpenAI's o1 series~\cite{openai2024o1}, agents can now perform deeper introspection and self-correction.
Furthermore, modern agents are increasingly adept at using external tools and APIs to execute code or retrieve information~\cite{wang2024survey,tran2501multi}.
Despite these advancements, most existing multi-agent systems focus on generating text or code in isolation. 
They cannot coordinate multiple modalities simultaneously—such as synchronizing a spoken script with dynamic visual animations—which is a critical requirement for high-quality educational content creation.

\paragraph{\textbf{Video Generation and Educational AI}}
Research in video generation can be categorized into two main directions: general-purpose video synthesis and domain-specific educational content creation. 
In the field of general synthesis, diffusion-based models have made significant progress in generating photorealistic videos from text prompts. Leading models, such as Sora~\cite{openai2024sora} and the open-source Wan 2.1~\cite{wan_etal_2025_wan}, use transformer architectures to maintain temporal consistency and visual quality~\cite{ho2022video,minimax2025hailuo,ding2025kling, runwayml2024introducing}. 
However, these models operate in ``pixel space,'' meaning they predict pixel values directly. This approach often lacks precise control over logical details, leading to factual errors or ``hallucinations'' that are unacceptable in mathematics or science education (e.g., drawing a triangle with four sides).
In the educational domain, AI applications have traditionally focused on personalized tutoring and ``talking head'' videos. 
Digital human platforms use Text-to-Speech (TTS) technology to drive lip-syncing for virtual avatars, reducing the cost of video production~\cite{raihan2025large,elai2024}. 
Additionally, Intelligent Tutoring Systems (ITS) provide adaptive feedback to students based on their performance~\cite{khanacademy2024khanmigo,cao2025survey}. 
While effective for engagement, these systems are mostly static or text-based. They rarely generate dynamic, explanatory visualizations that can evolve in real-time. 
Recent efforts in multimodal education have attempted to combine text and images~\cite{xu2024eduagent,zhang2024mathverse}, but they still struggle to produce logically accurate animations. 
Our work bridges this gap by using a multi-agent system to generate executable code (rather than pixels) for visualizations, ensuring that the instructional videos are both visually engaging and mathematically rigorous.

\paragraph{\textbf{Code-Centric Educational Video Generation}}
Recent video generation systems have begun to explore the use of LLM agents to produce executable animation scripts. 
For example, Code2Video~\cite{chen2025code2video} utilizes a tripartite pipeline (Planner $\to$ Coder $\to$ Critic) for direct text-to-code translation, while ANVIL~\cite{noviello2026anvilanalogiesvideoslecturers} proposes a multimodal system that generates analogy-based instructional animations for computer science topics by producing executable Manim code from concept definitions. Some of these works further emphasize student-centered evaluation, such as TeachQuiz-style knowledge transfer~\cite{chen2025code2video}. In contrast, \ourmethod is a multi-agent system built around a central Orchestrating Agent that jointly coordinates solution reasoning, executable illustration, narration, heterogeneous critique, and release filtering. This design targets industrial instructional-video production, where correctness, renderability, temporal alignment, and deployment gating are enforced holistically.
\section{Preliminary: From Probabilistic Pixels to Pedagogical Scripts}
\label{sec:preliminary}

To address the inherent limitations of end-to-end generative models, this section formalizes the transition from pixel-level approximation to a structured paradigm where instructional videos are treated as \textit{compiled pedagogical intents}. We argue that generating educational content is not a problem of predicting plausible pixels, but of orchestrating precise symbolic and temporal logic.

\subsection{Probabilistic Synthesis vs. Pedagogical Rigor}
Current text-to-video models primarily generate content by learning the distribution $p(\mathbf{X} | \mathbf{t})$, where $\mathbf{X}$ is a sequence of video frames and $\mathbf{t}$ is a text prompt. While this paradigm excels at producing visually coherent scenes, it is fundamentally misaligned with the requirements of instructional content, which prioritizes \textit{logical and symbolic fidelity} over aesthetic variance. 
For K--12 problem-solving scenarios, this misalignment leads to three critical failures:
\begin{itemize}
    \item \textbf{Symbolic Hallucination}: Mathematical equations and geometric diagrams are rendered inconsistently, as the model approximates their visual appearance rather than executing their semantic definitions.
    \item \textbf{Logical Non-sequiturs}: The causal, step-by-step progression of a solution is not explicitly modeled, resulting in contradictory reasoning or disjointed transitions between frames.
    \item \textbf{Temporal Semantic Drift}: The precise timing required for an explanation (audio) to coincide with its corresponding visual demonstration (e.g., highlighting a specific term) cannot be guaranteed by stochastic frame prediction.
\end{itemize}
These issues stem from a root cause: \textit{the output space (pixels) is disconnected from the domain's core semantic units (symbols and logical steps)}.

\subsection{Video as an Executable Script}
As aforementioned, instructional videos typically follow a structured pedagogical workflow that demands high logical density.
To this end, we propose to generate a structured \textit{instructional script} that can be deterministically compiled into a video. We define an \textbf{Executable Video Script (EVS)}, denoted as $\mathcal{S}$, as follows:
\begin{equation}
\label{eq:evs}
\mathcal{S} = (\mathcal{P}, \;\mathcal{N}, \;\mathcal{A}),
\end{equation}
where $\mathcal{P}$ (Pedagogical Content) is the modality-agnostic instructional core, including problem premises, step-by-step analytical reasoning, and definitions of visual assets (e.g., geometric constructs and LaTeX-rendered equations). $\mathcal{N}$ (Narration) is a temporal sequence of spoken explanations $\{n_1, n_2, \dots, n_k\}$ that verbalizes $\mathcal{P}$. Each segment $n_i$ corresponds to a discrete logical unit in the solution process. $\mathcal{A}$ (Alignment \& Orchestration) is a set of programmatic rules that synchronizes $\mathcal{P}$ and $\mathcal{N}$. This includes temporal mapping, stylistic consistency (fonts, colors), and the transition logic that governs scene evolution.

\subsection{Deterministic Video Synthesis}
The EVS is compiled into a final video $\mathcal{V}$ through a deterministic rendering pipeline:
\begin{equation}
\label{eq:compile}
\mathcal{V} = \texttt{Compile}(\mathcal{S}) = \texttt{Render}_{\text{vis}}(\mathcal{P}, \mathcal{A}) \parallel \texttt{Synth}_{\text{audio}}(\mathcal{N}),
\end{equation}
where $\texttt{Render}_{\text{vis}}$ is a programmatic rendering engine\footnote{In our implementation, we primarily utilize the Manim (Mathematical Animation) library to interpret $\mathcal{P}$ and $\mathcal{A}$ into precise visual streams.} that interprets the visual assets to produce a precise visual stream. $\texttt{Synth}_{\text{audio}}$ converts the narration $\mathcal{N}$ into speech. The $\parallel$ operator denotes their execution under the strict synchronization constraints defined in $\mathcal{A}$.
This paradigm ensures pedagogical rigor through:
\begin{enumerate}
    \item \textbf{Symbolic Fidelity}: Visuals are generated via \textbf{symbolic execution}\footnote{For instance, using \texttt{Write(Tex())} or \texttt{Create(Circle())} primitives ensures that equations and shapes are mathematically exact.} rather than pixel approximation.
    \item \textbf{Structural Consistency}: The solution's logical flow is explicitly encoded in the structure of $\mathcal{P}$.
    \item \textbf{Temporal Precision}: Synchronization is treated as a \textit{first-class citizen} in $\mathcal{A}$, enabling frame-accurate alignment between narration and visual events via dynamic synchronization primitives\footnote{For instance, the \texttt{self.wait(T)} primitive is dynamically calculated to bridge the gap between animation completion and narration end.}.
\end{enumerate}

\subsection{Generation as Script Synthesis}
The EVS formulation reframes the video generation task. The challenge shifts from \textit{``generating pixels that match a description''} to \textit{``synthesizing a coherent triplet $\mathcal{S} = (\mathcal{P}, \mathcal{N}, \mathcal{A})$ that fully specifies the intended pedagogy.''} This structured generation task naturally decomposes into specialized sub-tasks: content reasoning ($\mathcal{P}$), multimedia orchestration ($\mathcal{A}$), and pedagogical narration ($\mathcal{N}$), providing the foundational motivation for the multi-agent system introduced in Section~\ref{sec:methodology}.
\begin{figure*}
    \centering
    \includegraphics[width=0.95\linewidth]{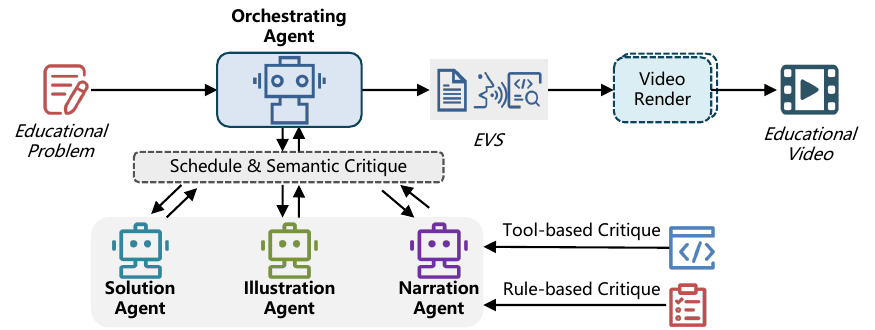}
    \caption{Overview of \ourmethod. A collaborative multi-agent harness centers on the Orchestrating Agent, which coordinates three specialized working agents--Solution Agent, Illustration Agent, and Narration Agent-- through iterative quality critique and template-based video assembly. The educational videos are thus rendered from assembled executable video scripts (EVS).}
    \Description{A system diagram showing an orchestrating agent coordinating solution, illustration, and narration agents, with semantic, tool-based, and rule-based critique loops before template-based video assembly.}
    \label{fig:lasev_overview}
\end{figure*}

\section{\ourmethod: The LLM-Based Multi-Agent System for Educational Video Generation}
\label{sec:methodology}

\subsection{System Overview}

The core objective of \ourmethod is to synthesize an \textbf{Executable Video Script (EVS)} (as defined in Eq.~\ref{eq:evs}) for an educational problem $Q$, and deterministically compile it into a final video. 

\paragraph{Collaborative multi-agent harness for EVS synthesis.}
To this end, \ourmethod synthesizes $\mathcal{S}$ through a collaborative multi-agent harness that integrates role-specialized generation, validation, and assembly.
Specifically, the harness centers on a single \textit{Orchestrating Agent} that analyzes the educational problem $Q$, maintains the shared production state, mediates inter-agent dependencies, consolidates validation feedback, and produces the alignment specification $\mathcal{A}$.
The \textit{Orchestrating Agent} coordinates three specialized working agents to generate the content components:
\begin{itemize}
    \item \textbf{Solution Agent} (\emph{solving}): generates rigorous step-by-step reasoning and structured pedagogical text ($\mathcal{P}^{\text{text}}$).
    \item \textbf{Illustration Agent} (\emph{illustrating}, optional): generates \emph{executable} visualization code (e.g., Python/Manim) and visual asset specifications ($\mathcal{P}^{\text{illus}}$).
    \item \textbf{Narration Agent} (\emph{narrating}): generates learner-friendly narration $\mathcal{N}$ aligned to the solution structure.
\end{itemize}
The harness follows a hierarchical control pattern: the orchestrator dispatches tasks to specialized agents, each agent refines its artifact under validation feedback, and validated outputs are recorded in the shared production state before EVS assembly.
Importantly, the pedagogical content $\mathcal{P}$ is the aggregation of outputs from the Solution and Illustration agents:
\begin{equation}
\mathcal{P} = \big[\;\mathcal{P}^{\text{text}} \cup \mathcal{P}^{\text{illus}}\;\big], \qquad \mathcal{P}^{\text{illus}} \ \text{is optional},
\end{equation}
where $\mathcal{P}^{\text{text}}$ is the modality-agnostic instructional core (premises, steps, symbolic elements), and $\mathcal{P}^{\text{illus}}$ contains executable visuals (code + asset definitions). This design explicitly implements the ``video as an executable script'' paradigm from Section~\ref{sec:preliminary}: the system does not directly synthesize pixels, but instead compiles validated symbolic content into deterministic visuals.

\subsection{Orchestrating Agent: Hierarchical Coordination}
The Orchestrating Agent serves as the control module of the collaborative multi-agent harness, maintaining a shared global state and enforcing cross-component consistency. Given an input problem $Q$, it synthesizes the EVS through:
\begin{equation}
\mathcal{S}=\mathcal{M}(Q) = \big(\mathcal{P},\mathcal{N},\mathcal{A}\big),
\end{equation}
where $\mathcal{M}$ denotes the Orchestrating Agent together with the working agents. Concretely, $\mathcal{M}$ performs (i) process planning and (ii) heterogeneous critique with iterative revision, and ultimately writes the final $\mathcal{A}$ that aligns content and narration for deterministic compilation.

\paragraph{Process planning.}
The Orchestrating Agent decomposes $Q$ into executable subtasks and decides whether optional illustration is beneficial. It dispatches structured task specifications to working agents while preserving pedagogical intent and target audience constraints, and records validated intermediate outputs so that downstream agents can condition on reliable prior artifacts.

\textit{Example (boat allocation).} Consider: ``A class of 46 students goes boating, taking 10 boats in total. Large boats hold 6 people, small boats hold 4 people, all fully occupied. How many large and small boats are there?'' The Orchestrating Agent identifies it as a classic constraint-solving problem. It instructs the Solution Agent to produce a stepwise reasoning trace suitable for K--12 explanation, and specifies that symbolic elements (e.g., $10\times4$, $46-40$) must be rendered deterministically. It then decides that dynamic illustration will meaningfully reduce cognitive load, scheduling the Illustration Agent to visualize the steps derived by the Solution Agent.

\paragraph{Heterogeneous critique and iterative revision.}
To ensure \textbf{logical and symbolic fidelity}, every intermediate artifact is gated by explicit validation. Let $o_i$ denote the output artifact produced at stage $i$. We define the orchestrator’s evaluation as:
\begin{align}
\mathrm{pass}_i &= \mathcal{C}_{\text{sem}}(o_i) \ \wedge\ \mathcal{C}_{\text{tool}}(o_i)\ \wedge\ \mathcal{C}_{\text{rule}}(o_i), \\
f_i &= \texttt{MergeFeedback}\big(f^{\text{sem}}_i, f^{\text{tool}}_i, f^{\text{rule}}_i\big),
\end{align}
where $\mathrm{pass}_i\in\{\text{True},\text{False}\}$ and $f_i$ is structured critique feedback. If $\mathrm{pass}_i=\text{False}$, the orchestrator returns $f_i$ to the corresponding agent for regeneration, forming a critique--revision loop until the quality gate is satisfied.
These checks serve as harness-level quality gates that determine whether each role output is reused, revised, or passed to downstream assembly.
This evaluation is \emph{heterogeneous} across three dimensions:
\begin{itemize}
    \item \textbf{Semantic Critique} ($\mathcal{C}_{\text{sem}}$): 
    LLM-based logical inspection of pedagogical content and reasoning chains based on predefined \textbf{rubrics} or curriculum-aligned syllabi, ensuring conceptual correctness, pedagogical coherence, and learner-friendliness. 
    In addition to rubric-based evaluation, \ourmethod incorporates carefully designed few-shot examples within the critique prompt to calibrate judgment criteria and enhance the stability and consistency of semantic assessment across iterative revisions.
    \item \textbf{Tool-based Critique} ($\mathcal{C}_{\text{tool}}$): Deterministic verification via external \textbf{executors or compilers} (e.g., Python/Manim); it captures hard errors and runtime exceptions to ensure visual assets are fully executable.
    \item \textbf{Rule-based Critique} ($\mathcal{C}_{\text{rule}}$): Automated auditing of structural compliance, including \textbf{keyword matching}, format validation, and adherence to specific function rules or API constraints.
\end{itemize}

\textit{Example (boat allocation).} For $\mathcal{P}^{\text{text}}$, the Orchestrating Agent checks the chain (assumption $10\times4=40$, discrepancy $46-40=6$) and verifies that the conclusion matches constraints. For $\mathcal{P}^{\text{illus}}$, the orchestrator first checks whether the illustration code faithfully reflects the validated steps and is executable; then, it uses tool-based critique (i.e., the Python interpreter) to capture execution errors and rule-based critique to ensure specific rendering functions are correctly invoked.

\paragraph{Constructing $\mathcal{A}$ (alignment \& orchestration).}
After $\mathcal{P}$ and $\mathcal{N}$ are validated, the Orchestrating Agent writes $\mathcal{A}$ as a set of programmatic rules that enforce:
(i) temporal mapping between narration segments and visual events,
(ii) style constraints (fonts, layout, transitions), which are mainly predefined in templates, and
(iii) scene evolution logic.
Crucially, $\mathcal{A}$ is \emph{not} produced by any working agent: it is derived globally by the orchestrator to guarantee end-to-end consistency under compilation.

\subsection{Working Agents: Domain-Specific Generation}
\ourmethod contains three working agents to specialize in generating complementary parts of the EVS. 
Each agent supports feedback-conditioned refinement based on its respective critique.

\paragraph{Solution Agent (rigorous solving, producing $\mathcal{P}^{\text{text}}$).}
Given $Q$ and feedback $f_{\mathcal{S}}$ from the Orchestrating Agent, the Solution Agent outputs structured pedagogical content:
\begin{equation}
\mathcal{P}^{\text{text}} = \mathcal{M}_{\text{sol}}(Q, f_{\mathcal{S}}).
\end{equation}
The output is organized as a sequence of logically atomic steps. It is primarily evaluated by \textbf{Semantic Critique} for logical rigor and \textbf{Rule-based Critique} for formatting compliance.

\paragraph{Illustration Agent (executable illustration, producing optional $\mathcal{P}^{\text{illus}}$).}
Conditioned on $Q$, validated $\mathcal{P}^{\text{text}}$, and optional feedback $f_{\mathcal{I}}$ from the Orchestrating Agent, it produces executable visualization assets:
\begin{equation}
\mathcal{P}^{\text{illus}} = \mathcal{M}_{\text{ill}}(Q, \mathcal{P}^{\text{text}}, f_{\mathcal{I}}).
\end{equation}
In our implementation, $\mathcal{P}^{\text{illus}}$ consists of Python/Manim code. The agent must resolve \textbf{Tool-based Critique} (compiler errors) and \textbf{Rule-based Critique} (function constraints) to ensure the code is renderable and stylistically consistent.

\begin{table*}[!thbp]
\centering
\caption{Performance comparison on Elementary Chinese Language Arts and Middle School Mathematics. \ourmethod achieves consistent improvements in Usable/Publishable Rate across both Chinese Language Arts and Mathematics problems.}
\label{tab:combined-results}
\begin{tabular}{llcccc}
\toprule
\textbf{Dataset} & \textbf{Method} & \textbf{Key Features} & \textbf{Usable Rate} & \textbf{Publishable Rate} & \textbf{Perfect Rate} \\
\midrule
\multicolumn{6}{l}{\textit{Elementary Chinese Language Arts}} \\
\midrule
\multirow{3}{*}{\shortstack[l]{Direct\\Prompting}} 
& GPT-4o & Single-model generation & 74.0 & 52.0 & 0.0 \\
& DeepSeek-R1 & Reinforcement learning & 80.0 & 70.0 & 5.0 \\
& Qwen3 & Chinese-optimized & 84.0 & 82.0 & 6.0 \\
\cmidrule{1-6}
& \textbf{\ourmethod} & \textbf{Full System} & \textbf{96.0} & \textbf{92.0} & 4.0 \\
\midrule
\midrule
\multicolumn{6}{l}{\textit{Middle School Mathematics}} \\
\midrule
\multirow{4}{*}{\shortstack[l]{Direct\\Prompting}}
& DeepSeek-R1 & Basic prompting & 45.5 & 45.5 & 27.3 \\
& DeepSeek-R1 & Enhanced prompting & 60.0 & 50.0 & 30.0 \\
& GPT-4o & Single-pass generation & 77.6 & 77.6 & 44.9 \\
& GPT-4o & Improved prompting & 94.0 & 86.0 & 52.0 \\
\cmidrule{1-6}
& \textbf{\ourmethod} & \textbf{Full System} & \textbf{96.0} & \textbf{96.0} & \textbf{58.0} \\
\bottomrule
\end{tabular}
\end{table*}

\paragraph{Narration Agent (pedagogical narration, producing $\mathcal{N}$).}
Given $Q$, validated $\mathcal{P}^{\text{text}}$ (and optionally $\mathcal{P}^{\text{illus}}$), and feedback $f_{\mathcal{N}}$ from the Orchestrating Agent, the Narration Agent generates a temporally segmentable narration:
\begin{equation}
\mathcal{N} = \mathcal{M}_{\text{nar}}(Q, \mathcal{P}^{\text{text}}, f_{\mathcal{N}}).
\end{equation}
It is governed by \textbf{Semantic Critique} to ensure the explanation aligns with the visual steps and \textbf{Rule-based Critique} to ensure the output can be correctly segmented for synchronization.

\subsection{Controllable Video Generation}
After constructing and validating $\mathcal{S}=(\mathcal{P},\mathcal{N},\mathcal{A})$, \ourmethod performs deterministic compilation as Eq.~\ref{eq:compile}.
We implement this via \textbf{template-driven assembly} controlled by $\mathcal{A}$. Templates define global visual identity and expose controllable slots for $\mathcal{P}$ and $\mathcal{N}$. The Orchestrating Agent populates these slots and writes explicit alignment rules, including:
(1) \textbf{Step-to-time mapping}: each narration segment $n_i$ is mapped to a visual event window.
(2) \textbf{Duration control}: display time is determined by symbolic density.
(3) \textbf{Optional illustration routing}: if $\mathcal{P}^{\text{illus}}$ is present, executable animation clips are inserted with synchronized triggers.
By replacing probabilistic generation with this critique-driven compilation-centric design, the harness turns collaboratively validated agent outputs into deterministic rendering inputs, and \ourmethod ensures that symbolic fidelity and temporal precision are strictly enforced\footnote{In practical deployments, \ourmethod may optionally integrate a \textit{digital human} into the video via an independent module. This part is outside the scope of this work.}.
\section{Experiments}

\subsection{Experimental Setup}

\begin{table*}[t]
\centering
\caption{Ablation study validating each system component. Multi-agent collaboration provides the largest contribution, followed by Semantic Critique, Tool-based Critique, and Rule-based Critique.}
\label{tab:ablation}
\begin{tabular}{lcccccc}
\toprule
\multirow{2}{*}{\textbf{Configuration}} & \multicolumn{3}{c}{\textbf{Chinese Language Arts}} & \multicolumn{3}{c}{\textbf{Mathematics}} \\
\cmidrule(lr){2-4} \cmidrule(lr){5-7}
& \textbf{Usable Rate} & \textbf{Publishable Rate} & \textbf{Perfect Rate} & \textbf{Usable Rate} & \textbf{Publishable Rate} & \textbf{Perfect Rate} \\
\midrule
Single-agent only & 64.0 & 42.0 & 0.0 & 60.0 & 50.0 & 30.0 \\
w/o Semantic Critique & 74.0 & 52.0 & 0.0 & 77.6 & 77.6 & 48.0 \\
\ \ \ \ w/o Few-shot Examples  & 90.0 & 86.0 & 6.0 & 84.0 & 84.0 & 30.0 \\
w/o Tool-based Critique & 86.0 & 69.8 & 2.3 & 78.0 & 60.0 & 16.0 \\
w/o Rule-based Critique & 84.0 & 82.0 & 10.0 & 88.0 & 80.0 & 16.0 \\
\midrule
\textbf{Full System (\ourmethod)} & \textbf{96.0} & \textbf{92.0} & 4.0 & \textbf{96.0} & \textbf{96.0} & \textbf{58.0} \\
\bottomrule
\end{tabular}
\end{table*}

\paragraph{Datasets}
We evaluate our system on two representative K--12 educational datasets. 
(1) \textbf{Elementary Chinese Language Arts} consists of text-based reading comprehension and sentence analysis problems, which emphasize semantic understanding, structural parsing, and pedagogical explanation.
(2) \textbf{Middle School Mathematics} covers algebra, geometry, and word problems, including both text-only and figure-accompanied questions, and requires precise symbolic reasoning and step-by-step solution presentation.
Together, these datasets reflect complementary challenges in educational video generation across language understanding and mathematical reasoning.
The offline evaluation set contains 493 Chinese Language Arts problems and 510 Mathematics problems, and each problem is paired with one generated instructional video for expert assessment.

\paragraph{Evaluation Metrics}
All generated videos are evaluated by domain experts using a 4-level rubric:
\textbf{Score 3} (perfect correctness with professional-level presentation),
\textbf{Score 2} (conceptually correct with minor presentation or clarity issues),
\textbf{Score 1} (noticeable flaws but still pedagogically usable),
and \textbf{Score 0} (incorrect or unsuitable for instructional use).
Based on this rubric, we report three aggregated metrics:
\textbf{Usable Rate} (the percentage of videos with score $\geq 1$),
\textbf{Publishable Rate} (the percentage of videos with score $\geq 2$, serving as the primary metric for real-world deployment),
and \textbf{Perfect Rate} (the percentage of videos receiving score $=3$).
To ensure evaluation reliability, each generated video is independently assessed by three certified domain experts with at least one year of teaching experience. Scores are aggregated by majority voting, with senior expert arbitration for complete disagreement; all three experts assign identical scores to 85\% of the evaluated videos.

\paragraph{Baselines}
We compare \ourmethod against strong direct-prompting baselines built on state-of-the-art LLMs, which represent the prevailing practice for educational content generation without explicit multi-agent coordination or iterative quality control.
For the Chinese dataset, we evaluate GPT-4o (single-pass generation)~\cite{achiam2023gpt}, DeepSeek-R1 (reinforcement learning–enhanced reasoning)~\cite{guo2025deepseek}, and Qwen3 (Chinese-optimized training)~\cite{yang2025qwen3}.
For the Mathematics dataset, we test DeepSeek-R1 under both basic and enhanced prompting strategies, as well as GPT-4o with single-pass and improved prompting.
These baselines isolate the performance of powerful foundation models when used in a single-stage generation setting, allowing us to quantify the gains introduced by structured multi-agent collaboration and critique-driven refinement.

\paragraph{Implementation}
We use DeepSeek-R1 as the backbone LLM with a maximum of 3 refinement iterations per stage. The Illustration Agent uses Manim for visualizations with tool-based and rule-based critique. 
Videos are assembled with structured templates and text-to-speech synthesis.

\subsection{Main Results}
\begin{table*}[t]
\centering
\caption{Cost comparison across video generation approaches. Our \ourmethod achieves around 95\% cost reduction over education industry baselines while maintaining 96\% quality.}
\label{tab:cost-comparison}
\resizebox{\textwidth}{!}{
\begin{tabular}{llcccc}
\toprule
\textbf{Category} & \textbf{Method} & \textbf{Cost/Video} & \textbf{Usable Rate} & \textbf{Capacity (/day)} & \textbf{Trade-off} \\
\midrule
\multirow{3}{*}{\shortstack[l]{Traditional\\Education}} 
& Industry Standard & \$1.40 (10 CNY) & 95 & 30K & Quality + Scale $\rightarrow$ High Cost \\
& Baidu Vertical (1-year opt.) & \$0.90 (6.4 CNY) & 88 & 500 & Quality + Cost $\rightarrow$ Low Scale \\
& VideoTutor (AI-based) & \$0.90--1.33 & 45 & Low & None optimized \\
\midrule
\multirow{6}{*}{\shortstack[l]{Video\\Generation}} 
& Runway Gen-3 & \$1.00 (10s) & -- & High & Cost + Scale $\rightarrow$ Low Quality \\
& Sora 2 Standard & \$1.00 (10s) & -- & Medium & Cost + Scale $\rightarrow$ Low Quality \\
& Sora 2 Pro & \$5.00 (10s) & -- & Low & Quality only $\rightarrow$ High Cost \\
& Kling AI & \$0.05--0.49 (5s) & 26.7 & High & Cost + Scale $\rightarrow$ Low Quality \\
& Hailuo AI & \$0.45 (12s) & -- & High & Cost + Scale $\rightarrow$ Low Quality \\
& Vidu & \$0.22 (6s) & -- & High & Cost + Scale $\rightarrow$ Low Quality \\
\midrule
\multirow{3}{*}{\ourmethod} 
& LLM (DeepSeek-R1) & \$0.056 (0.4 CNY) & -- & -- & -- \\
& Digital Human & $\leq$\$0.014 (0.1 CNY) & -- & -- & -- \\
& \textbf{Full System (2-min video)} & \textbf{$\leq$\$0.07(0.5 CNY)} & \textbf{96} & \textbf{1M} & \textbf{All Three Balanced} \\
\bottomrule
\end{tabular}
}
\end{table*}

\begin{table}[t]
\centering
\caption{Error distribution in Math final outputs. }
\label{tab:error-analysis}
\begin{tabular}{lr}
\toprule
\textbf{Error Type} & \textbf{Math (\%)} \\
\midrule
\multicolumn{2}{l}{\textit{Critical Issues (Score $\leq 1$)}} \\
\quad Solution Error & 2.0 \\
\quad Misreading Problem & 2.0 \\
\midrule
\multicolumn{2}{l}{\textit{Minor Issues (Score = $2$)}} \\
\quad Presentation flaws & 22.0 \\
\quad Imprecise description & 4.0 \\
\quad Minor redundancy & 6.0 \\
\quad Minor step skipping & 2.0 \\
\quad Non-optimal method & 2.0 \\
\quad Auxiliary line issues & 2.0 \\
\midrule
\textbf{Critical Issue Rate} & \textbf{4.0} \\
\textbf{Minor Issue Rate} & \textbf{38.0} \\
\textbf{Total Non-Perfect} & \textbf{42.0} \\
\bottomrule
\end{tabular}
\end{table}

Table~\ref{tab:combined-results} shows the performance comparison between our \ourmethod and baseline methods across both datasets. 
On Elementary Chinese Language Arts, state-of-the-art LLMs achieve Publishable Rate scores ranging from 52.0\% to 82.0\%, with Qwen3 performing best at 82.0\% due to its Chinese-specific training. DeepSeek-R1 reaches 70.0\% through reinforcement learning, while GPT-4o achieves 52.0\%. While the rule-based critique reduces Perfect Rate from 10.0\% to 4.0\% on Chinese, it is essential for achieving a high Publishable Rate by catching edge-case errors.
On Middle School Mathematics, direct prompting methods show Usable Rate scores between 45.5\% and 94.0\%, with optimized GPT-4o reaching the highest baseline performance. 
However, these single-pass approaches lack quality-control mechanisms, as evidenced by their Publishable Rate and Perfect Rate scores on the math dataset, making them less suitable for real-world use despite strong initial results. 
In contrast, our full system achieves 96.0\% Usable Rate and 92.0\% Publishable Rate on Chinese, representing a 10-40 percentage point improvement over baselines. 
On Mathematics, \ourmethod also achieves state-of-the-art results (96.0\% Usable Rate, 96.0\% Publishable Rate, and notably 58.0\% Perfect Rate).
Compared to baseline models, \ourmethod ensures consistent quality through specialized working agents, quality evaluation, and iterative refinement.
Real-world large-scale deployment validates that our approach achieves production-ready quality at scale, with 92\%--96\% of generated videos passing expert review for deployment (score $\geq 2$).

\subsection{Ablation Studies}

Table~\ref{tab:ablation} shows the contribution of each component. The single-agent baseline achieves only 42.0\% Publishable Rate on Chinese and 60.0\% Usable Rate on Math, confirming that single-model generation without multi-agent decomposition and quality control is inadequate for educational video generation. Semantic Critique ($\mathcal{C}_{\text{sem}}$) provides the largest gain (40.0 points on Chinese Publishable Rate: 52.0\% $\rightarrow$ 92.0\%), demonstrating that LLM-based logical inspection with rubric-guided evaluation is essential for pedagogical correctness; within it, few-shot examples further stabilize evaluation consistency by providing concrete quality standards. Tool-based Critique ($\mathcal{C}_{\text{tool}}$) contributes 22.2 points (Chinese Publishable Rate) by catching Python runtime errors and Manim compilation failures through external executors. 
Rule-based Critique adds 10.0 points by checking format compliance and keyword requirements.
The full system reaches a 58.0\% Perfect Rate on Math, and removing any single component reduces this by at least 10 percentage points, confirming that logical reasoning, executable verification, and format checking must work together to ensure symbolic correctness. 
However, on Chinese, the Full System achieves a lower Perfect Rate (4.0\%) than some ablated variants (e.g., 10.0\% without Rule-based Critique) because strict structural compliance triggers revisions that elevate borderline cases from Score~1 to Score~2 (Publishable Rate: 82.0\%~$\rightarrow$~92.0\%), but occasionally introduce minor rephrasing artifacts scored as 2 rather than 3—a \emph{desirable} trade-off as the primary deployment metric improves substantially.

\begin{figure*}[htbp]
\centering
\begin{tabular}{|c|c|c|}
\hline
\textbf{Subject} & \textbf{Positive Example} & \textbf{Negative Example} \\
\hline
\textbf{Chinese} & 
\includegraphics[width=0.42\textwidth,valign=c]{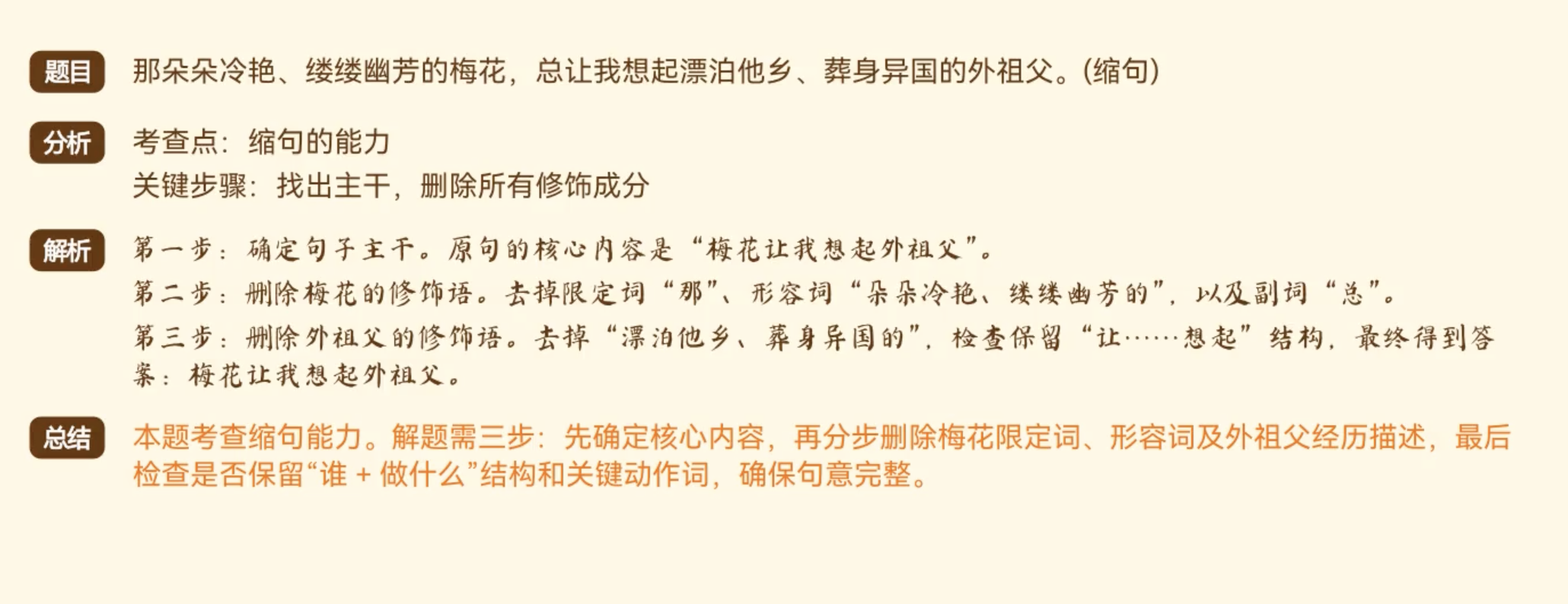} & 
\includegraphics[width=0.42\textwidth,valign=c]{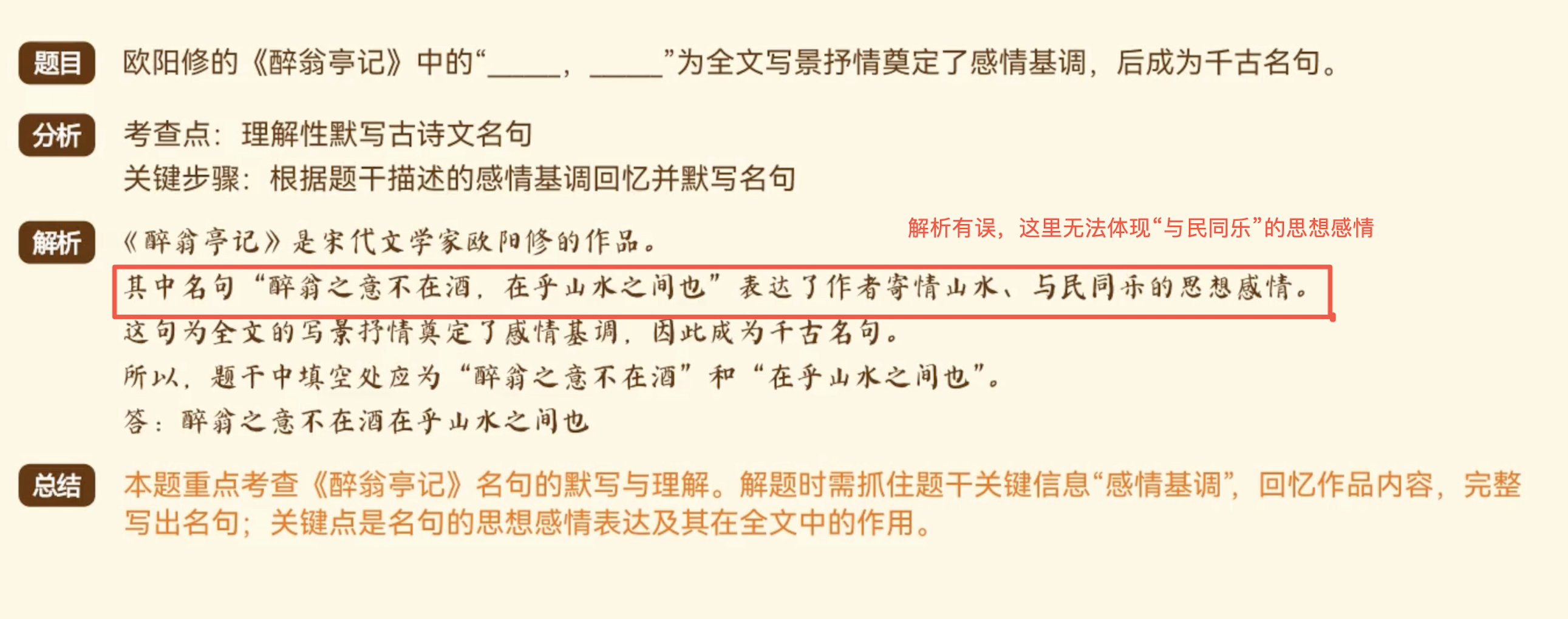} \\
\hline
\textbf{Math} & 
\includegraphics[width=0.42\textwidth,valign=c]{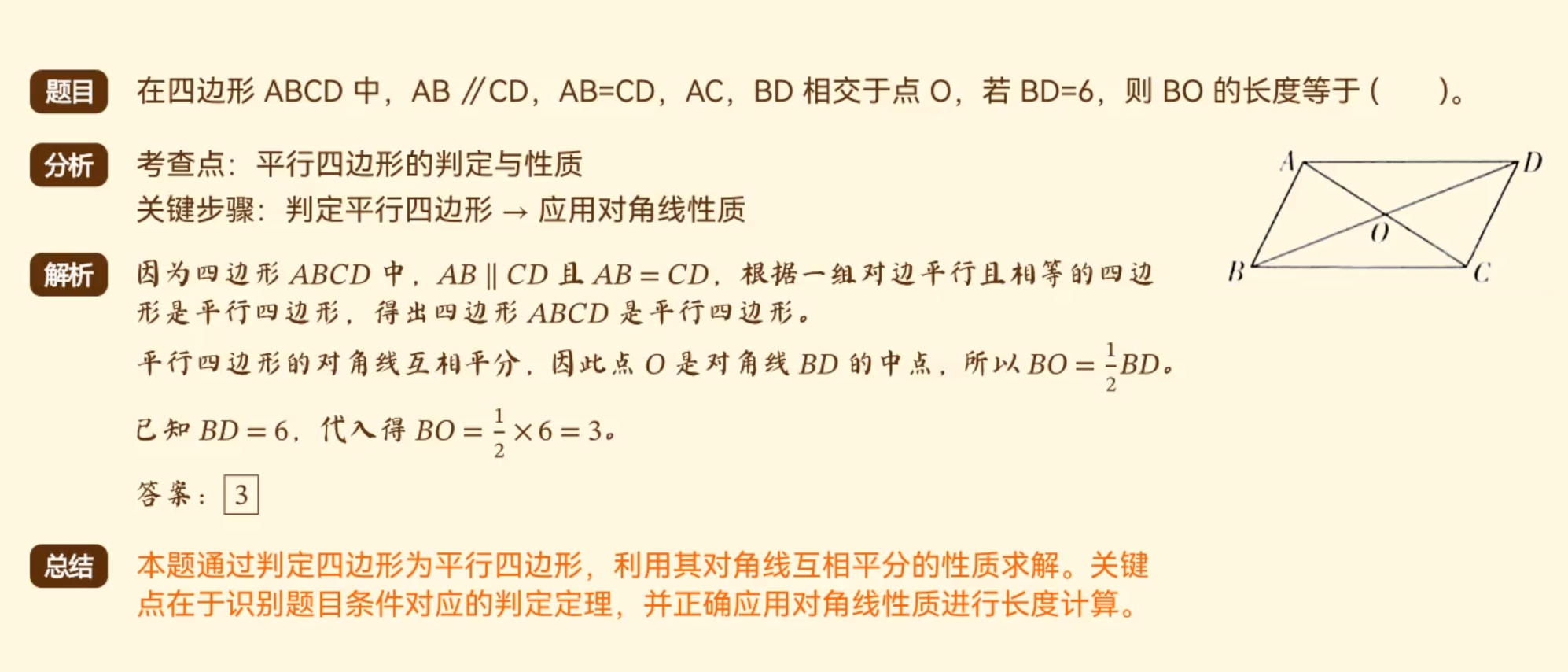} & 
\includegraphics[width=0.42\textwidth,valign=c]{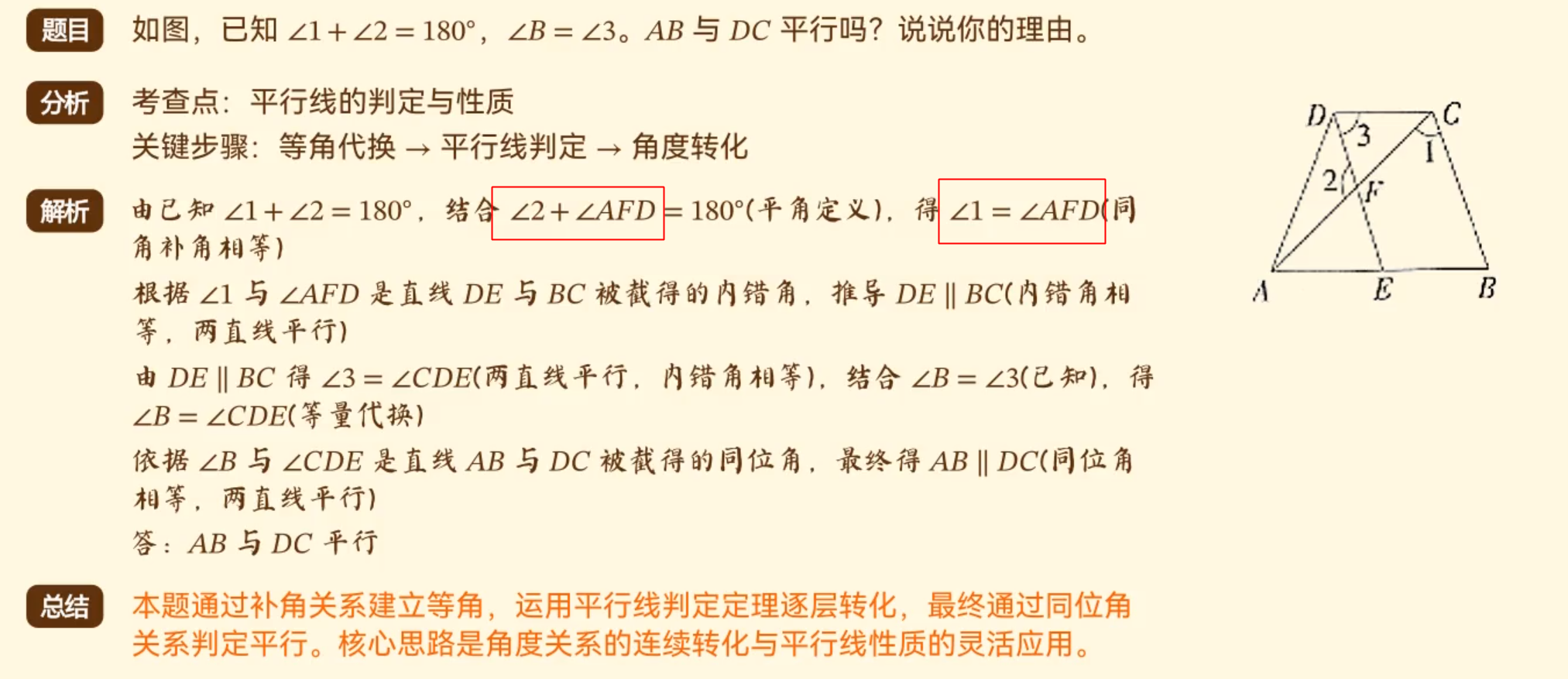} \\
\hline
\end{tabular}
\caption{Several positive and negative cases produced by \ourmethod across real Chinese Language Arts and Mathematics problems.}
\Description{A comparison table with positive and negative examples for Chinese Language Arts and Mathematics, showing screenshots of generated instructional video frames.}
\label{fig:case_study}
\end{figure*}

\subsection{Error Analysis}

Table~\ref{tab:error-analysis} reveals the error distribution in Math videos, showing that most issues (38.0\%) are minor presentation issues scored as 2, such as presentation flaws (22.0\%), minor redundancy (6.0\%), and imprecise description (4.0\%) that do not affect core teaching quality, while critical issues account for only 4.0\% with solution errors at 2.0\% and problem misreading at 2.0\%. The non-perfect rate (42.0\%) primarily reflects these minor presentation issues rather than serious teaching problems, as evidenced by the low critical issue rate (4.0\%) and the high Perfect Rate (58.0\%), showing that over half of the videos need no revision at all. This result confirms that our three-stage critique mechanism effectively prevents major failures: Semantic Critique ($\mathcal{C}_{\text{sem}}$) catches logical errors and misreadings through rubric-based evaluation, Tool-based Critique ($\mathcal{C}_{\text{tool}}$) ensures computational correctness via Python execution, and Rule-based Critique ($\mathcal{C}_{\text{rule}}$) addresses format compliance including presentation standards and notation requirements, together keeping critical issues below 5\% while achieving production-ready quality for educational deployment.

\paragraph{Large-Scale Deployment Performance}
The quality-control procedure described above is also applied during online deployment. When a module fails semantic, tool-based, or rule-based validation, the corresponding module is regenerated with at most three refinement attempts per stage; candidates that still fail these quality gates are filtered before release. Production logs covering over 20 million generated videos show that the overall failure-or-filtering rate is 3.58\%. As shown in Table~\ref{tab:deployment-failure}, over 96\% of videos are successfully produced, while fewer than 4\% are filtered before release. In absolute terms, policy-based filtering accounts for 1.30\% of all production cases, and the two main technical failure modes---solution/script generation and animation/LaTeX compilation---account for 1.32\% and 0.90\%, respectively. This suggests that residual failures arise mainly from bounded local generation and tool-execution issues, rather than from unstable iterative drift.

\begin{table}[t]
\centering
\caption{Production outcome distribution in large-scale online deployment.}
\label{tab:deployment-failure}
\begin{tabular}{@{}lr@{}}
\toprule
\textbf{Outcome Category} & \textbf{Percentage (\%)} \\
\midrule
Successful Production & 96.42 \\
\rowcolor{gray!20}
\textit{Failed Production} & \textit{3.58} \\
\hspace{2em}Policy-based filtering & 1.30 \\
\hspace{2em}Solution/script generation failure & 1.32 \\
\hspace{2em}Animation/LaTeX compilation failure & 0.90 \\
\hspace{2em}Infrastructure exception & 0.06 \\
\bottomrule
\end{tabular}
\end{table}

\subsection{Cost and Scalability Analysis}

Table~\ref{tab:cost-comparison} compares our \ourmethod against mainstream approaches. Traditional education companies achieve 95\% quality through expert annotation but incur a \$1.40 per-video cost at a 30K daily capacity. Baidu's optimized vertical reduces cost to \$0.90 after one year of development, achieving 88\% quality but limiting throughput to 500 videos per day due to human review bottlenecks. Commercial video generation models (Runway~\cite{runwayml2024introducing}, Sora~\cite{openai2024sora}, Kling~\cite{ding2025kling}, Hailuo~\cite{minimax2025hailuo}, Vidu~\cite{bao2024vidu}) offer \$0.05--5.00 pricing with high throughput but achieve only 26.7\% quality for educational content, as they lack domain knowledge for mathematical notation, pedagogical structure, and solution correctness. Our \ourmethod achieves less than \$0.07 per 2-minute video (\$0.056 for DeepSeek-R1 and less than \$0.014 for digital human), representing a 95\% cost reduction compared to traditional industry-standard costs while maintaining 96\% quality.
Existing methods face a quality-cost-scale trilemma where optimizing two dimensions sacrifices the third: traditional education optimizes quality and scale but incurs high cost (\$1.40), commercial AI optimizes cost and scale but achieves low quality (26.7--45\%), and Baidu's vertical optimizes quality and cost but limits scale (500/day). \ourmethod uniquely balances all three through its hierarchical multi-agent coordination: domain-specific agents provide automated quality assurance, achieving 96\% without human intervention, template-based generation with few-shot examples reduces LLM calls for 5$\times$ cost reduction, and a fully automated workflow enables 1M videos per day, limited only by API throughput. At scale, covering China's 5M K12 problems requires 5 days and \$0.35 Million investment versus 167 days and \$7M for traditional methods.

\subsection{Case Study}

Figure~\ref{fig:case_study} illustrates representative cases demonstrating the effectiveness and the remaining limitations of our \ourmethod. 
The positive cases demonstrate that our collaborative multi-agent harness effectively guides the base model to produce high-quality outputs. 
In Chinese, the system generates a well-structured three-step solution: 
(1) identifying the core structure ``\chinese{梅花让我想起外祖父}'' (plum blossoms remind me of my maternal grandfather), 
(2) systematically removing modifiers for both subject and object with explicit rules, and 
(3) verifying structural preservation—demonstrating that our Narration Agent effectively produces well-structured pedagogical narration.
In Math, the solution exhibits rigorous two-stage reasoning by first invoking the parallelogram criterion (``one pair of opposite sides both parallel and equal''), then applying diagonal bisection to calculate $BO = \frac{1}{2}BD = 3$. 
These successes validate that our Rule-based Critique and Semantic Critique effectively steer the base model toward accurate and pedagogically sound outputs.
However, the negative cases expose inherent limitations of the base model. 
In Chinese, the base model incorrectly attributes ``\chinese{醉翁之意不在酒，在乎山水之间也}'' (The Drunken Old Man’s interest does not lie in the wine, but in the mountains and waters) to the ideal of ``\chinese{与民同乐}'' (enjoying with the people) rather than personal nature appreciation—a critical issue, reflecting insufficient literary knowledge in the base model's training corpus. 
In Math, the base model produces flawed geometric reasoning, claiming $\angle 2 = \angle AFD$ instead of the correct relationship $\angle 2 = \angle 3 + \angle DCF$—a solution error (2.0\%) that indicates gaps in the base model's logical inference capabilities for multi-step geometric proofs. 
Despite these limitations, \ourmethod achieves low critical issue rates (4.0\% Math) and high Perfect Rate (58.0\% Math), demonstrating that our quality control mechanisms effectively mitigate most base model weaknesses.
\section{Conclusion}
\label{sec:conclusion}

We propose a hierarchical LLM-based multi-agent system for industrial-scale instructional video generation. 
By breaking down video synthesis into specialized tasks—rigorous reasoning, pedagogical narration, and executable visualization—coordinated under centralized orchestration with heterogeneous critiques, our system addresses key limitations of end-to-end video generation models in handling symbolic content and procedural accuracy. 
Industrial deployment demonstrates the practical effectiveness at large scale, with a production capacity exceeding one million videos per day and over 95\% cost reduction compared to industry baselines, while maintaining an approximately 96\% Publishable Rate.
These results demonstrate that structured multi-agent collaboration can effectively transform instructional video generation from a labor-intensive bottleneck into a scalable, automated process. 
Overall, \ourmethod shows that a collaborative multi-agent harness can turn LLM reasoning and code generation into a deployable video-production workflow with explicit verification and release control.
The system can be extended to broader educational domains through domain-specific templates and validation rules. Future work will assess this generalization on public cross-disciplinary benchmarks and incorporate student-centered learning-effectiveness measures beyond expert review.

\begin{acks}
The authors would like to thank Zhifang Xue, Ruiyu Zhao, Zesong Zhang for their tremendous support during the course of this work. We also thank our colleagues at Baidu for their continuous support and for creating an environment conducive to applied research.
\end{acks}

\bibliographystyle{ACM-Reference-Format}
\balance
\bibliography{main}

@article{achiam2023gpt,
  title={Gpt-4 technical report},
  author={Achiam, Josh and Adler, Steven and Agarwal, Sandhini and Ahmad, Lama and Akkaya, Ilge and Aleman, Florencia Leoni and Almeida, Diogo and Altenschmidt, Janko and Altman, Sam and Anadkat, Shyamal and others},
  journal={arXiv preprint arXiv:2303.08774},
  year={2023}
}

@article{touvron2023llama,
  title={Llama: Open and efficient foundation language models},
  author={Touvron, Hugo and Lavril, Thibaut and Izacard, Gautier and Martinet, Xavier and Lachaux, Marie-Anne and Lacroix, Timoth{\'e}e and Rozi{\`e}re, Baptiste and Goyal, Naman and Hambro, Eric and Azhar, Faisal and others},
  journal={arXiv preprint arXiv:2302.13971},
  year={2023}
}

@article{zhao2023survey,
  title={A survey of large language models},
  author={Zhao, Wayne Xin and Zhou, Kun and Li, Junyi and Tang, Tianyi and Wang, Xiaolei and Hou, Yupeng and Min, Yingqian and Zhang, Beichen and Zhang, Junjie and Dong, Zican and others},
  journal={arXiv preprint arXiv:2303.18223},
  volume={1},
  number={2},
  year={2023}
}

@inproceedings{hong2023metagpt,
  title={MetaGPT: Meta programming for a multi-agent collaborative framework},
  author={Hong, Sirui and Zhuge, Mingchen and Chen, Jonathan and Zheng, Xiawu and Cheng, Yuheng and Wang, Jinlin and Zhang, Ceyao and Wang, Zili and Yau, Steven Ka Shing and Lin, Zijuan and others},
  booktitle={The twelfth international conference on learning representations},
  year={2023}
}

@article{li2023camel,
  title={Camel: Communicative agents for" mind" exploration of large language model society},
  author={Li, Guohao and Hammoud, Hasan and Itani, Hani and Khizbullin, Dmitrii and Ghanem, Bernard},
  journal={Advances in Neural Information Processing Systems},
  volume={36},
  pages={51991--52008},
  year={2023}
}

@article{qian2023communicative,
  title={Communicative agents for software development},
  author={Qian, Chen and Cong, Xin and Yang, Cheng and Chen, Weize and Su, Yusheng and Xu, Juyuan and Liu, Zhiyuan and Sun, Maosong},
  journal={arXiv preprint arXiv:2307.07924},
  volume={6},
  number={3},
  pages={1},
  year={2023}
}

@inproceedings{chen2023agentverse,
  title={Agentverse: Facilitating multi-agent collaboration and exploring emergent behaviors},
  author={Chen, Weize and Su, Yusheng and Zuo, Jingwei and Yang, Cheng and Yuan, Chenfei and Chan, Chi-Min and Yu, Heyang and Lu, Yaxi and Hung, Yi-Hsin and Qian, Chen and others},
  booktitle={The Twelfth International Conference on Learning Representations},
  year={2023}
}

@inproceedings{park2023generative,
  title={Generative agents: Interactive simulacra of human behavior},
  author={Park, Joon Sung and O'Brien, Joseph and Cai, Carrie Jun and Morris, Meredith Ringel and Liang, Percy and Bernstein, Michael S},
  booktitle={Proceedings of the 36th annual acm symposium on user interface software and technology},
  pages={1--22},
  year={2023}
}

@article{bo2024reflective,
  title={Reflective multi-agent collaboration based on large language models},
  author={Bo, Xiaohe and Zhang, Zeyu and Dai, Quanyu and Feng, Xueyang and Wang, Lei and Li, Rui and Chen, Xu and Wen, Ji-Rong},
  journal={Advances in Neural Information Processing Systems},
  volume={37},
  pages={138595--138631},
  year={2024}
}

@article{shinn2023reflexion,
  title={Reflexion: Language agents with verbal reinforcement learning},
  author={Shinn, Noah and Cassano, Federico and Gopinath, Ashwin and Narasimhan, Karthik and Yao, Shunyu},
  journal={Advances in Neural Information Processing Systems},
  volume={36},
  pages={8634--8652},
  year={2023}
}

@article{wang2024survey,
  title={A survey on large language model based autonomous agents},
  author={Wang, Lei and Ma, Chen and Feng, Xueyang and Zhang, Zeyu and Yang, Hao and Zhang, Jingsen and Chen, Zhiyuan and Tang, Jiakai and Chen, Xu and Lin, Yankai and others},
  journal={Frontiers of Computer Science},
  volume={18},
  number={6},
  pages={186345},
  year={2024},
  publisher={Springer}
}

@misc{openai2024o1,
  title={Learning to Reason with LLMs},
  author={OpenAI},
  year={2024},
  howpublished={\url{https://openai.com/index/learning-to-reason-with-llms/}}
}

@misc{openai2024sora,
  title={Sora: Creating Video from Text},
  author={OpenAI},
  year={2024},
  howpublished={\url{https://openai.com/sora}}
}

@article{ho2022video,
  title={Video diffusion models},
  author={Ho, Jonathan and Salimans, Tim and Gritsenko, Alexey and Chan, William and Norouzi, Mohammad and Fleet, David J},
  journal={Advances in neural information processing systems},
  volume={35},
  pages={8633--8646},
  year={2022}
}

@inproceedings{raihan2025large,
  title={Large language models in computer science education: A systematic literature review},
  author={Raihan, Nishat and Siddiq, Mohammed Latif and Santos, Joanna CS and Zampieri, Marcos},
  booktitle={Proceedings of the 56th ACM Technical Symposium on Computer Science Education V. 1},
  pages={938--944},
  year={2025}
}

@article{xu2024eduagent,
  title={Eduagent: Generative student agents in learning},
  author={Xu, Songlin and Zhang, Xinyu and Qin, Lianhui},
  journal={arXiv preprint arXiv:2404.07963},
  year={2024}
}

@inproceedings{zhang2024mathverse,
  title={Mathverse: Does your multi-modal llm truly see the diagrams in visual math problems?},
  author={Zhang, Renrui and Jiang, Dongzhi and Zhang, Yichi and Lin, Haokun and Guo, Ziyu and Qiu, Pengshuo and Zhou, Aojun and Lu, Pan and Chang, Kai-Wei and Qiao, Yu and others},
  booktitle={European Conference on Computer Vision},
  pages={169--186},
  year={2024},
  organization={Springer}
}

@misc{elai2024,
  title={Elai.io Automated Video Platform},
  author={Elai},
  year={2024},
  howpublished={\url{https://elai.io}}
}

@misc{khanacademy2024khanmigo,
  title={Khanmigo: AI-Powered Tutor},
  author={Khan Academy},
  year={2024},
  howpublished={\url{https://www.khanacademy.org/khan-labs}}
}

@article{guo2025deepseek,
  title={Deepseek-r1: Incentivizing reasoning capability in llms via reinforcement learning},
  author={Guo, Daya and Yang, Dejian and Zhang, Haowei and Song, Junxiao and Zhang, Ruoyu and Xu, Runxin and Zhu, Qihao and Ma, Shirong and Wang, Peiyi and Bi, Xiao and others},
  journal={arXiv preprint arXiv:2501.12948},
  year={2025}
}

@article{tran2501multi,
  title={Multi-agent collaboration mechanisms: A survey of LLMs, 2025},
  author={Tran, Khanh-Tung and Dao, Dung and Nguyen, Minh-Duong and Pham, Quoc-Viet and O’Sullivan, Barry and Nguyen, Hoang D},
  journal={URL https://arxiv. org/abs/2501.06322}
}

@article{cao2025survey,
  title={A survey of ai-generated content (aigc)},
  author={Cao, Yihan and Li, Siyu and Liu, Yixin and Yan, Zhiling and Dai, Yutong and Yu, Philip and Sun, Lichao},
  journal={ACM Computing Surveys},
  volume={57},
  number={5},
  pages={1--38},
  year={2025},
  publisher={ACM New York, NY}
}

@misc{jiang2024morahighrankupdatingparameterefficient,
      title={MoRA: High-Rank Updating for Parameter-Efficient Fine-Tuning}, 
      author={Ting Jiang and Shaohan Huang and Shengyue Luo and Zihan Zhang and Haizhen Huang and Furu Wei and Weiwei Deng and Feng Sun and Qi Zhang and Deqing Wang and Fuzhen Zhuang},
      year={2024},
      eprint={2405.12130},
      archivePrefix={arXiv},
      primaryClass={cs.CL},
      url={https://arxiv.org/abs/2405.12130}, 
}

@article{wang2025mavis,
  title={MAViS: A multi-agent framework for long-sequence video storytelling},
  author={Wang, Qian and Huang, Ziqi and Jia, Ruoxi and Debevec, Paul and Yu, Ning},
  journal={arXiv preprint arXiv:2508.08487},
  year={2025}
}

@misc{ghafoorian2026rehyatrecurrenthybridattention,
      title={ReHyAt: Recurrent Hybrid Attention for Video Diffusion Transformers}, 
      author={Mohsen Ghafoorian and Amirhossein Habibian},
      year={2026},
      eprint={2601.04342},
      archivePrefix={arXiv},
      primaryClass={cs.CV},
      url={https://arxiv.org/abs/2601.04342}, 
}

@inproceedings{wang2025emcontrol,
  title={EMControl: Adding Conditional Control to Text-to-Image Diffusion Models via Expectation-Maximization},
  author={Wang, He and Dai, Longquan and Tang, Jinhui},
  booktitle={Proceedings of the AAAI Conference on Artificial Intelligence},
  volume={39},
  number={7},
  pages={7691--7699},
  year={2025}
}

@article{wang2025lavie,
  title={Lavie: High-quality video generation with cascaded latent diffusion models},
  author={Wang, Yaohui and Chen, Xinyuan and Ma, Xin and Zhou, Shangchen and Huang, Ziqi and Wang, Yi and Yang, Ceyuan and He, Yinan and Yu, Jiashuo and Yang, Peiqing and others},
  journal={International Journal of Computer Vision},
  volume={133},
  number={5},
  pages={3059--3078},
  year={2025},
  publisher={Springer}
}

@article{tang2025video,
  title={Video understanding with large language models: A survey},
  author={Tang, Yunlong and Bi, Jing and Xu, Siting and Song, Luchuan and Liang, Susan and Wang, Teng and Zhang, Daoan and An, Jie and Lin, Jingyang and Zhu, Rongyi and others},
  journal={IEEE Transactions on Circuits and Systems for Video Technology},
  year={2025},
  publisher={IEEE}
}

@misc{gao_etal_2025_seedance_10,
  title = {Seedance 1.0: {{Exploring}} the {{Boundaries}} of {{Video Generation Models}}},
  shorttitle = {Seedance 1.0},
  author = {Gao, Yu and Guo, Haoyuan and Hoang, Tuyen and Huang, Weilin and Jiang, Lu and Kong, Fangyuan and Li, Huixia and Li, Jiashi and Li, Liang and Li, Xiaojie and Li, Xunsong and Li, Yifu and Lin, Shanchuan and Lin, Zhijie and Liu, Jiawei and Liu, Shu and Nie, Xiaonan and Qing, Zhiwu and Ren, Yuxi and Sun, Li and Tian, Zhi and Wang, Rui and Wang, Sen and Wei, Guoqiang and Wu, Guohong and Wu, Jie and Xia, Ruiqi and Xiao, Fei and Xiao, Xuefeng and Yan, Jiangqiao and Yang, Ceyuan and Yang, Jianchao and Yang, Runkai and Yang, Tao and Yang, Yihang and Ye, Zilyu and Zeng, Xuejiao and Zeng, Yan and Zhang, Heng and Zhao, Yang and Zheng, Xiaozheng and Zhu, Peihao and Zou, Jiaxin and Zuo, Feilong},
  year = 2025,
  month = jun,
  number = {arXiv:2506.09113},
  eprint = {2506.09113},
  primaryclass = {cs},
  publisher = {arXiv},
  doi = {10.48550/arXiv.2506.09113},
  urldate = {2026-02-02},
  archiveprefix = {arXiv}
}

@misc{seedance_etal_2025_seedance_15_pro,
  title = {Seedance 1.5 pro: {{A Native Audio-Visual Joint Generation Foundation Model}}},
  shorttitle = {Seedance 1.5 Pro},
  author = {Seedance Team},
  year = 2025,
  month = dec,
  number = {arXiv:2512.13507},
  eprint = {2512.13507},
  primaryclass = {cs},
  publisher = {arXiv},
  doi = {10.48550/arXiv.2512.13507},
  urldate = {2026-02-02},
  archiveprefix = {arXiv}
}

@misc{wan_etal_2025_wan,
  title = {Wan: {{Open}} and {{Advanced Large-Scale Video Generative Models}}},
  shorttitle = {Wan},
  author = {Wan Team},
  year = 2025,
  month = apr,
  number = {arXiv:2503.20314},
  eprint = {2503.20314},
  primaryclass = {cs},
  publisher = {arXiv},
  doi = {10.48550/arXiv.2503.20314},
  urldate = {2026-02-02},
  archiveprefix = {arXiv}
}

@article{rosenshine2012principles,
  title={Principles of instruction: Research-based strategies that all teachers should know.},
  author={Rosenshine, Barak},
  journal={American educator},
  volume={36},
  number={1},
  pages={12},
  year={2012},
  publisher={ERIC}
}

@book{mayer2021multimedia,
  title={Multimedia Learning},
  author={Mayer, Richard E},
  year={2021},
  publisher={Cambridge University Press}
}

@book{anderson2001taxonomy,
  title={A taxonomy for learning, teaching, and assessing: A revision of Bloom's taxonomy of educational objectives: complete edition},
  author={Anderson, Lorin W and Krathwohl, David R},
  year={2001},
  publisher={Addison Wesley Longman, Inc.}
}

@incollection{sweller2011cognitive,
  title={Cognitive load theory},
  author={Sweller, John},
  booktitle={Psychology of learning and motivation},
  volume={55},
  pages={37--76},
  year={2011},
  publisher={Elsevier}
}

@inproceedings{wu2024autogen,
title={AutoGen: Enabling Next-Gen {LLM} Applications via Multi-Agent Conversations},
author={Qingyun Wu and Gagan Bansal and Jieyu Zhang and Yiran Wu and Beibin Li and Erkang Zhu and Li Jiang and Xiaoyun Zhang and Shaokun Zhang and Jiale Liu and Ahmed Hassan Awadallah and Ryen W White and Doug Burger and Chi Wang},
booktitle={First Conference on Language Modeling},
year={2024},
url={https://openreview.net/forum?id=BAakY1hNKS}
}

@inproceedings{chen2025improving,
title={Improving Retrieval-Augmented Generation through Multi-Agent Reinforcement Learning},
author={Yiqun Chen and Lingyong Yan and Weiwei Sun and Xinyu Ma and Yi Zhang and Shuaiqiang Wang and Dawei Yin and Yiming Yang and Jiaxin Mao},
booktitle={The Thirty-ninth Annual Conference on Neural Information Processing Systems},
year={2025},
url={https://openreview.net/forum?id=9Ia0KiVAut}
}

@misc{yan_etal_2025_beyond,
  title = {Beyond {{Self-Talk}}: {{A Communication-Centric Survey}} of {{LLM-Based Multi-Agent Systems}}},
  shorttitle = {Beyond {{Self-Talk}}},
  author = {Yan, Bingyu and Zhou, Zhibo and Zhang, Litian and Zhang, Lian and Zhou, Ziyi and Miao, Dezhuang and Li, Zhoujun and Li, Chaozhuo and Zhang, Xiaoming},
  year = 2025,
  month = jun,
  number = {arXiv:2502.14321},
  eprint = {2502.14321},
  primaryclass = {cs},
  publisher = {arXiv},
  doi = {10.48550/arXiv.2502.14321}
}

@misc{tran_etal_2025_multiagent,
  title = {Multi-{{Agent Collaboration Mechanisms}}: {{A Survey}} of {{LLMs}}},
  shorttitle = {Multi-{{Agent Collaboration Mechanisms}}},
  author = {Tran, Khanh-Tung and Dao, Dung and Nguyen, Minh-Duong and Pham, Quoc-Viet and O'Sullivan, Barry and Nguyen, Hoang D.},
  year = 2025,
  month = jan,
  number = {arXiv:2501.06322},
  eprint = {2501.06322},
  primaryclass = {cs},
  publisher = {arXiv},
  doi = {10.48550/arXiv.2501.06322}
}

@misc{li_etal_2025_a_unified,
  title = {A {{Unified Multi-Agent Framework}} for {{Universal Multimodal Understanding}} and {{Generation}}},
  author = {Li, Jiulin and Huang, Ping and Li, Yexin and Chen, Shuo and Hu, Juewen and Tian, Ye},
  year = 2025,
  month = aug,
  number = {arXiv:2508.10494},
  eprint = {2508.10494},
  primaryclass = {cs},
  publisher = {arXiv},
  doi = {10.48550/arXiv.2508.10494}
}

@article{yang2025qwen3,
  title={Qwen3 technical report},
  author={Yang, An and Li, Anfeng and Yang, Baosong and Zhang, Beichen and Hui, Binyuan and Zheng, Bo and Yu, Bowen and Gao, Chang and Huang, Chengen and Lv, Chenxu and others},
  journal={arXiv preprint arXiv:2505.09388},
  year={2025}
}

@article{bao2024vidu,
  title={Vidu: a highly consistent, dynamic and skilled text-to-video generator with diffusion models},
  author={Bao, Fan and Xiang, Chendong and Yue, Gang and He, Guande and Zhu, Hongzhou and Zheng, Kaiwen and Zhao, Min and Liu, Shilong and Wang, Yaole and Zhu, Jun},
  journal={arXiv preprint arXiv:2405.04233},
  year={2024}
}

@article{ding2025kling,
  title={Kling-avatar: Grounding multimodal instructions for cascaded long-duration avatar animation synthesis},
  author={Ding, Yikang and Liu, Jiwen and Zhang, Wenyuan and Wang, Zekun and Hu, Wentao and Cui, Liyuan and Lao, Mingming and Shao, Yingchao and Liu, Hui and Li, Xiaohan and others},
  journal={arXiv preprint arXiv:2509.09595},
  year={2025}
}

@techreport{minimax2025hailuo,
  title={MiniMax-01: Scaling Foundation Models with Lightning Attention},
  author={MiniMax},
  year={2025},
  institution={MiniMax},
  url={https://github.com/MiniMax-AI}
}

@misc{runwayml2024introducing,
  title={Introducing gen-3 alpha: a new frontier for video generation},
  author={RunwayML, GA},
  year={2024},
  publisher={Runway Research}
}

@article{chen2025code2video,
  title={Code2Video: A Code-centric Paradigm for Educational Video Generation},
  author={Chen, Yanzhe and Lin, Kevin Qinghong and Shou, Mike Zheng},
  journal={arXiv preprint arXiv:2510.01174},
  year={2025}
}

@misc{noviello2026anvilanalogiesvideoslecturers,
      title={ANVIL: Analogies and Videos for Lecturers}, 
      author={Yuri Noviello and Anastasiia Birillo and Gosia Migut},
      year={2026},
      eprint={2605.16295},
      archivePrefix={arXiv},
      primaryClass={cs.CY},
      url={https://arxiv.org/abs/2605.16295}, 
}


\end{document}